\documentclass[letterpaper, 10 pt, conference]{ieeeconf}  

\IEEEoverridecommandlockouts                              
\usepackage{graphicx}
\usepackage{svg}
\usepackage{amsmath} 
\usepackage{amssymb}  

\usepackage{microtype}
\usepackage{hyperref}
\usepackage{textcomp}
\usepackage[ruled,vlined,linesnumbered]{algorithm2e}
\SetNlSty{texttt}{}{.}
\usepackage{tabulary}
\usepackage{multirow}
\usepackage{bm}
\usepackage{caption}
\usepackage{dblfloatfix}

\title{\LARGE \bf
Walking on TacTip toes:  A tactile sensing foot for walking robots} 

\author{Elizabeth A.\ Stone\textsuperscript{1}, Nathan F.\ Lepora\textsuperscript{2} and David A.W.\ Barton\textsuperscript{3}
\thanks{\textsuperscript{1}ES is a PhD student at the EPSRC Centre for Doctoral Training in Future Autonomous and Robotic Systems (FARSCOPE) at the Bristol Robotics Laboratory. \texttt{\small lizzie.stone@brl.ac.uk}}%
\thanks{\textsuperscript{2}NL  is with the Dept.\ of Engineering Maths and Bristol Robotics Laboratory, University of Bristol, UK, and is supported by the Leverhulme Trust (RL-2016-39). \texttt{\small n.lepora@bristol.ac.uk}}
\thanks{\textsuperscript{3}DB is with the Dept.\ of Engineering Maths, University of Bristol, UK.
        \texttt{\small david.barton@bristol.ac.uk}}%
}

\begin{document}

\maketitle
\thispagestyle{empty}
\pagestyle{empty}

\begin{abstract}
    Little research into tactile feet has been done for walking robots despite the benefits such feedback could give when walking on uneven terrain. This paper describes the development of a simple, robust and inexpensive tactile foot for legged robots based on a high-resolution biomimetic TacTip tactile sensor. Several design improvements were made to facilitate tactile sensing while walking, including the use of phosphorescent markers to remove the need for internal LED lighting. The usefulness of the foot is verified on a quadrupedal robot performing a beam walking task and it is found the sensor prevents the robot falling off the beam. Further, this capability also enables the robot to walk along the edge of a curved table. This tactile foot design can be easily modified for use with any legged robot, including much larger walking robots, enabling stable walking in challenging terrain.

\end{abstract}

\section{INTRODUCTION}
    Walking requires adaptation to the current terrain to successfully remain upright. Feedback from the environment is required to adapt to these changes in the terrain. Tactile sensors make contact with and interact directly with the environment and give high spacial resolution, giving insight into multiple properties of the contacted surface.  Wu et al.~\cite{Wu2019} recently showed that even a low resolution tactile sensor placed on weg style legs is enough to reliably classify the terrain to allow a robot to use terrain specific gaits to increase locomotion speed.
    
    Walking robots are currently limited in their use of direct feedback from the environment, relying only on a mixture of remote sensing methods (e.g. vision, sonar, lidar)~\cite{Meng2016}, force feedback sensing in joints and single force sensors on the feet to detect the terrain.  Remote methods are passive and are therefore limited to measurements that do not require interaction (for example, the compliance of surfaces cannot be measured, nor can the shape of a surface be found when obscured by ground cover), which is limiting especially in natural terrain where the ground is sometimes obscured by features such as long grass, leaves or snow. Force sensors on the other hand do make direct contact with the ground but do not provide high spatial resolution, often giving only a single dimension, which is not enough for complex analyses, such as efficient texture classification, edge detection or slip detection. Sensing in the joints (used to create virtual force sensors~\cite{Gonzalez2006}) suffers from the same limitations as direct force sensors.
    
    Tactile sensors on the other hand can measure a variety of properties; for example, the TacTip biomimetic optical tactile sensor~\cite{Ward-Cherrier2018} can measure tangential force, shear force, surface texture~\cite{pestell2018}, surface slip~\cite{James2018}, surface shape (in 3D)~\cite{Cramphorn2018} and  material compliance amongst others. Adding these modes of feedback may be beneficial for walking robots and enable stable walking in challenging terrain. 
    
    This paper takes the first steps in using tactile feet to aid walking of a robot in uneven terrain, demonstrated by the robot walking along the edge of a raised path without falling. It is found that the tactile foot alone is able to guide the robot along a safe path in challenging terrain, following the edge of a raised path and placing the foot at a safe distance from it. Adaptation to the terrain was implemented with an online learning algorithm to rapidly build a sensor model to interpret tactile stimuli~\cite{stone2019learning}.
    As far as the authors could find, this is the first demonstration of a walking robot with non-rotary legs being fitted with a high-resolution tactile-sensing foot. This appears also to be the first demonstration of a walking robot following-edge features using only tactile feedback via the feet.
    
    \begin{figure}[t]
        \centering
        
        \begin{tabular}[b]{@{}c@{}}
                \includegraphics[height=.21\textheight]{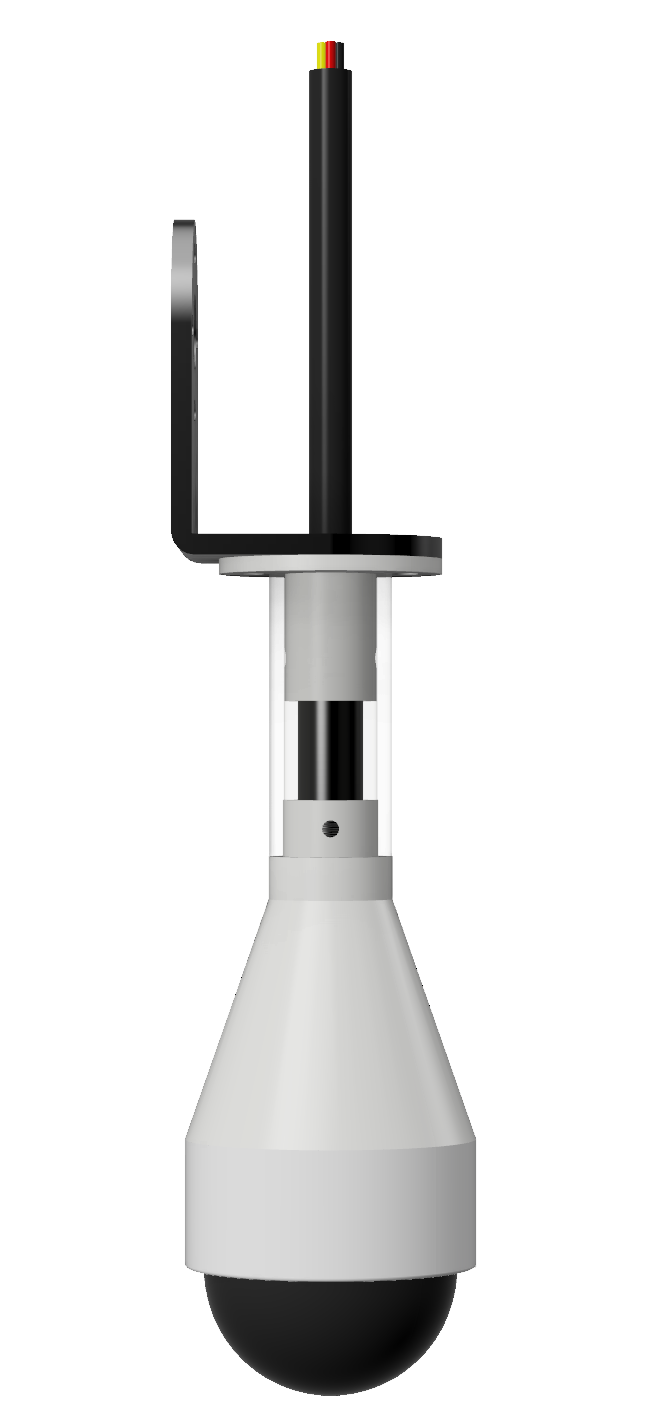}
                \includegraphics[height=.21\textheight]{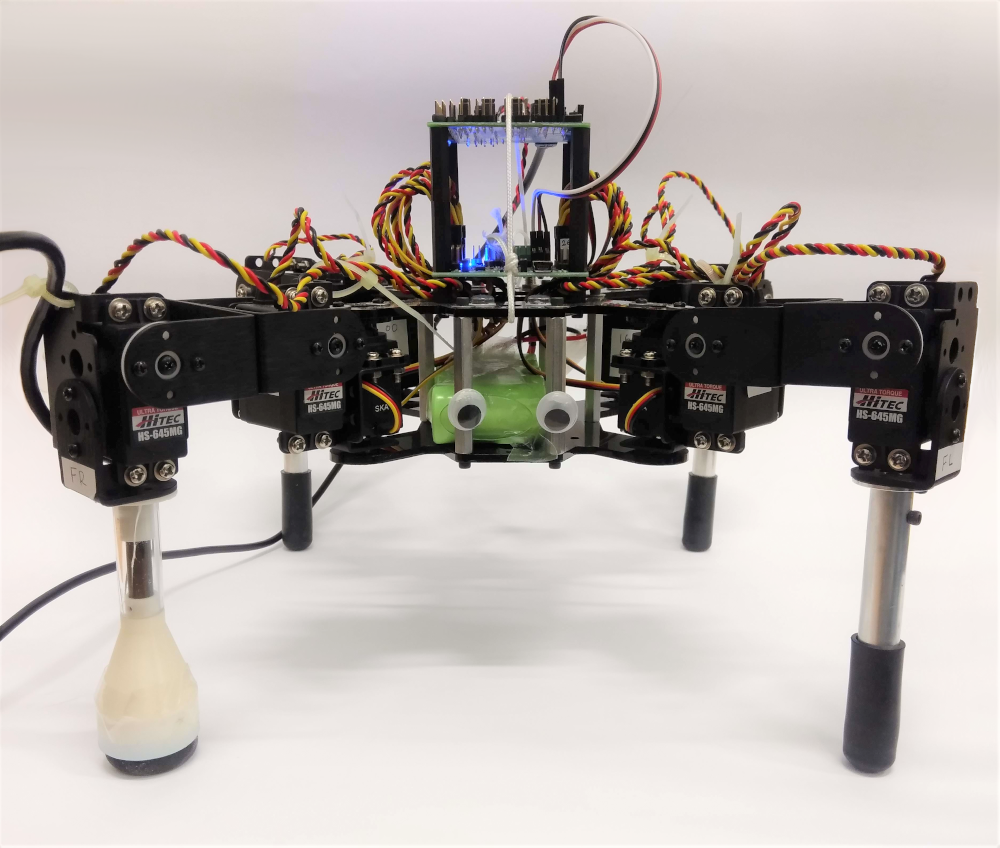} 
        \end{tabular}
    
        \caption{Left: CAD model of tactile foot. Right: Quadrupedal robot with front right foot replaced with tactile foot.}
        \label{robot}
        \vspace{-15pt}
    \end{figure}

\section{Background and Related Work}
    Previous attempts at creating tactile feet for walking robots have been few and far between and focused on texture classification. Recently, Wu et al.~\cite{Wu2019} developed small tactile sensors for weg style legs on small robots. They focused on texture classification to adapt gait to improve efficiency over different terrain. The sensor was only 6 taxels resolution (including one exclusively measuring the shear force) and therefore may not be capable of detecting fine features accurately (e.g. the exact angle of an edge). Despite the robot weighing less than 500g, the sensor could apparently function with up to 100N of force applied to it, implying that the technology could scale to larger robots.
    
    Previously Shill et al.~\cite{Shill2015} used a high-resolution pressure sensing array to classify terrain with a single detached foot and on a one legged hopping robot. The sensor was inspired by human neurons, implemented with piezo-electric strain gauges.  They found that, while the sensor was extremely accurate at identifying the correct terrain, sensor failure was a significant problem to the extent that the focus of the paper is in developing algorithms to deal with the sensor failure during experiments. The life span of their sensors is reported to be about 8 minutes without use of fault tolerant algorithms, increasing to 53 minutes with the use of repair filters.
    
    Overall it is likely that sensor fragility, cost of manufacture, and perceived complexities in system integration~\cite{Gonzalez2006} are the reasons that no commercial robots and only two research platforms using tactile feet could be found. The TacTip resolves these issues: the sensor is as robust as the gel inside, meaning very tough TacTips have been developed~\cite{Elkington2020}, and the sensor is rapidly and cheaply 3D printed, enabling quick and easy testing of different morphologies for specialist applications~\cite{Ward-Cherrier2018}. The sensor has also been integrated into various systems, from robot hands~\cite{church2019}~\cite{Ward-Cherrier2017} to tactile whiskered robots~\cite{Lepora2018a}.
    
    Instead of tactile sensors, walking robots often make use of other sensory feedback methods~\cite{Nobili2017}.
    The use of force feedback to balance is common~\cite{Silva2012}~\cite{Hardarson2002}~\cite{Focchi2018}~\cite{wieber2016modeling} and is particularly useful in bipedal robots~\cite{Gupta2011}~\cite{Kalamdani2007a}~\cite{Claros2014}~\cite{nelson2019petman} which are not statically stable so must rely on sensor feedback to remain upright. The center of pressure can be estimated using force sensors or virtual force sensors and can be extended to purposes beyond balance, for example, estimating the locations of edges and allowing bipedal robots to balance on these edges  \cite{Wiedebach2016} or to avoid them when climbing stairs \cite{Lee2017}. 
    
    It has has been argued that  sensors in the feet are redundant~\cite{Gonzalez2006}, adding additional complexity and points of failure, when virtual force sensors making use of servo feedback are sufficient and reliable for stable walking. This may be true for simple force sensors, but a tactile sensor gives many more dimensions and information than a force sensor, which cannot be sensed via the joints. In humans, the sense of touch in our feet is essential for walking, with impairments to this sensory system requiring the use of physical aids to walk. As such, it is not the aim to replace virtual force feedback systems, which work well for estimating balance and identifying the presence of contact in limbs. Instead tactile sensors can provide additional information that cannot be obtained through other methods, supplementing current methods of control and solving the remaining problems for walking robots in complex terrains~\cite{Li2011}.

\section{Methods}

    \subsection{Quadruped Robot}
\begin{figure}[t]
              \centering

              \includegraphics[width=.98\columnwidth]{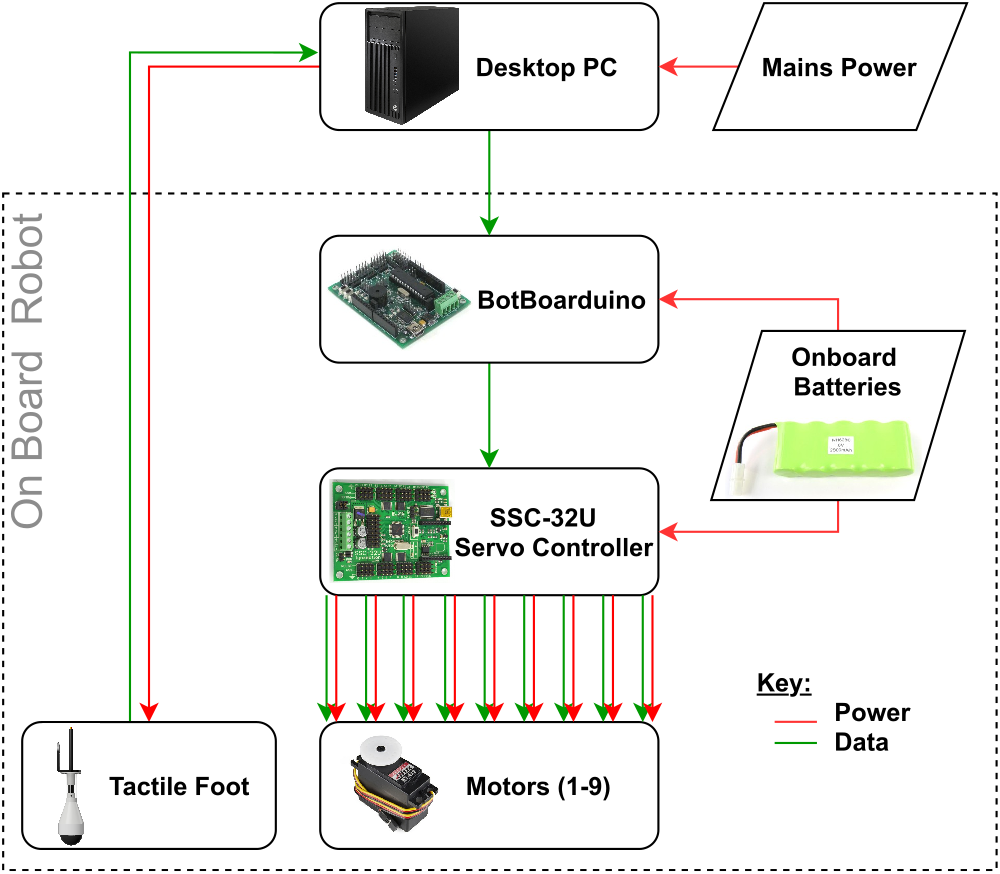}

              \caption{Diagram showing different components of the setup and how they interact with each other.}
              \label{data-diag}
        \end{figure}

        The robot used for this work was a Lynxmotion SQ3U \cite{RobotShop}, a 3DOF per leg quadruped robot. 
        The robot is symmetric across two axes and so is able to move side to side in addition to forward and back. It is able to achieve static creep gaits. It weighs approximately 1.5kg including on-board batteries, measures 23cm between adjacent feet and stands at 24cm tall (to the top of the boards, 16cm to the top of the legs) when upright as in \autoref{robot}. As seen in \autoref{data-diag}, the robot is Controlled by ``Botboarduino'', a modified version of the Aruduino Demulionove board, and a SSC-32U servo controller. The Botboarduino is attached via USB to the desktop PC which runs higher-level, more computationally-intense algorithms. The on-board battery powers the motors and boards, but the tactile sensor is powered instead via USB from the PC. In future, the use of a single-board computer (such as a Raspberry Pi) would remove all tethers and enable full autonomy.
        
        The robot was purchased as a build-it-yourself kit in a modular design that allowed easy addition of a tactile foot. The only modifications made to original robot were to swap the normal end-link for a tactile foot (TacFoot) and to extend other feet by 15mm (as the tactile foot is longer than the default foot). In this initial study, only one foot is replaced with a TacFoot to simplify the problem, in particular control, but the design would work for all feet on the robot. 
        
        \begin{figure*}[th]
              \centering
              \includegraphics[width=.98\textwidth]{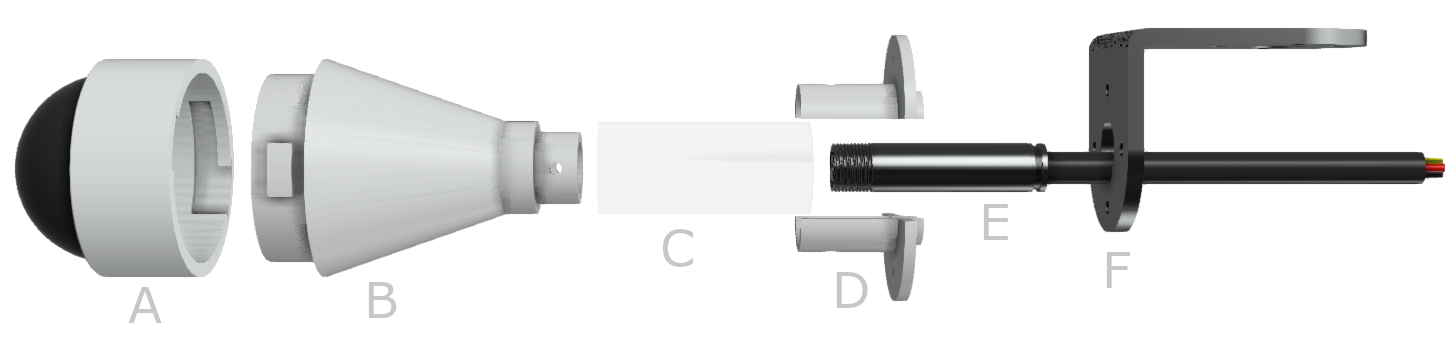}
              
              \caption{Exploded view of tactile foot CAD with A) Mini Tactip, B) foot cone, C) acrylic tube, D) split joint, E) endoscopic camera and F) robot's unmodified end leg bracket.}
              \label{exploded}
        \end{figure*}
        
    \subsection{Software}
        The robot came with no in-built code except the motor-controller commands (allowing the Arduino to request a motor number, position and time which the motor-controller then handles). All other code was implemented from scratch or modified from existing used with the TacTip.
        
        The desktop runs a similar software stack to~\cite{stone2019learning}; namely, the high-level algorithms are implemented in MATLAB, communicating with the tactile sensor through Python, which processes the images and passes an array of pin locations back to MATLAB. The difference here is that MATLAB communicates directly with the robot (where, in other studies, there was a layer of IronPython to communicate with a robotic arm). The MATLAB communicates with the  Arduino C code running on the BotBoarduino, sending high-level pose commands (e.g. \texttt{BR\_leg\_forward} would be an instruction to move the back-right leg forward). The C code simply contains a list of preset sequences required to walk in a straight line on a flat surface (inspired by \cite{Gonzalez2006}), and a function to rotate each leg about the hip to allow turning. The C code works out the series of commands to be sent to the motors to obtain this pose, and passes this on to the motor controller which drives the motors in the desired sequence. 
        
    \subsection{The TacTip} 
        The TacTip sensor is a biomimetic optical tactile sensor developed at Bristol Robotics Laboratory~\cite{chorley2009}~\cite{Ward-Cherrier2018} that consists of a domed, black rubber-like membrane (Tango Black+) with white-tipped pins (Vero White) protruding from the inner surface. The dome is filled with a transparent gel (Techsil RTV27905) and covered by an acrylic lens. A camera is placed on the other side of the lens to track the movements of the white pin tips. With any tactile stimulation, the pins deflect in distinctive ways, creating unique pin patterns that can be interpreted through statistical methods or machine learning algorithms. 
        
        The black membrane, gel and lens are referred to as the ``tip", which is easily detachable from the rest of the sensor, consisting of the camera, case to hold the tip and camera and attachments for mounting to platform, which are referred to collectively as the ``casing". This modular design allows easy experimentation with different tips, and replacement of broken tips, and use of the same tip on different cases and platforms. 
        
        As the TacTip is 3D printed, it has been developed into many shapes and sizes to preform many different tasks~\cite{Ward-Cherrier2018}. The majority of use cases have been as single sensors mounted on robotic arms, and in a variety of robotic hands. Of these there is only one case where the sensor been required to withstand application of significant weight, and that was mounted on a robotic arm for automatic application carbon fibre layup~\cite{Elkington2020}. To achieve sufficient resilience the normal gel was replaced with a much tougher material which retains its shape better under compression.

    \subsection{Modifications for the TacFoot} 
        The final design of the tactile foot can be seen in \autoref{robot} and a piece-by-piece breakdown of components can be seen in \autoref{exploded}.
        
        In line with the original TacTip, the foot has a modular design to allow easy updating and replacement of components, an ideal feature for a foot that in natural terrain may encounter hazardous features, such as sharp edges. The sensor is also designed such that the delicate and expensive parts (i.e. the camera) are kept away from damaging impacts, meaning only the inexpensive and easily manufactured parts (i.e. the tip) are likely to need replacing.
        
    \subsubsection{Casing Design}
        As previously mentioned the TacTip is normally mounted on a robotic arm or in fingers of robotic hands. With a robotic arm there are few limitations on size and weight of the TacTip casing, and little need for a power efficient design (as they are readily connected to the mains power supply). With robotic fingertips there are limitations on the size of the casing, needing to keep added bulk behind the tip to a minimum. These represent only very large and bulky designs, or very small and specialist designs -- here an in-between design is proposed, making use of all the available space a leg shape allows whilst remaining lightweight, cheap and simple to manufacture. The casing is made from 3D printed ABS and is 31mm wide (the full length of the sensor is 95mm).
        
        The casing for the foot is extremely simple, consisting of only 4 unique parts:
            
        \begin{itemize}
            \item Inspection/Endoscopic style 640x480 USB camera with 6 in-built white LEDs (with manually controlled light intensity) and a 5m cable.
            \item Cone to hold camera in line with tip, and to bear weight of robot. The cone shape maximises camera field of view while minimising size and weight.
            \item Two identical joint pieces which together connect the sensor to the robot's leg bracket with nuts and bolts. The two pieces together resembles the original joint piece but is modified to slot the camera inside and is split in two to allow the cable to thread through the robot's leg bracket (without needing to alter the cable).
            \item A tube to connect the cone to the joint and to protect the camera.
        \end{itemize}{}
        
    \subsubsection{Tip Choice}
        For the tip, a mini-TacTip~\cite{Pestell2018a} is used as this design has been verified before, and is a good compromise between being small enough for the robot to carry and being big enough to have a large tactile sensing area (tip diameter is 27mm at base of black dome). The hemispherical design was chosen here, instead of a flatter design, because, whilst the hemisphere makes less contact with a flat surface when the leg is perpendicular to the surface, the hemisphere retains roughly the same size contact area with surfaces rotated up to approximately 45\textdegree. As the robot will investigate a surface with an outstretched leg before putting weight through it in its normal stance, the sensor will often not be perpendicular to the surface when in use, therefore the design best able to sense across a variety of angles was chosen.
        
        The gel used inside the tip is the same as used for the normal TacTip~\cite{Ward-Cherrier2018}. As this gel is compliant, the sensor is compressed by the weight of the robot and gives a very strong signal -- this sensitivity could be decreased  by using a stiffer gel, depending on the application.
      
    \subsubsection{Novel Lighting}
        Novel lighting was investigated to allow simpler design and easier manufacture of the TacFoot, aid pin detection and potentially improve power consumption of the sensor.
        
\begin{figure}[t]
              \centering
              
              \begin{tabular}[b]{@{}cc@{}}
                    {\bf \small Normal pins} & {\bf \small Phosphorescent pins} \\[0pt]
                    \includegraphics[width=.47\columnwidth]{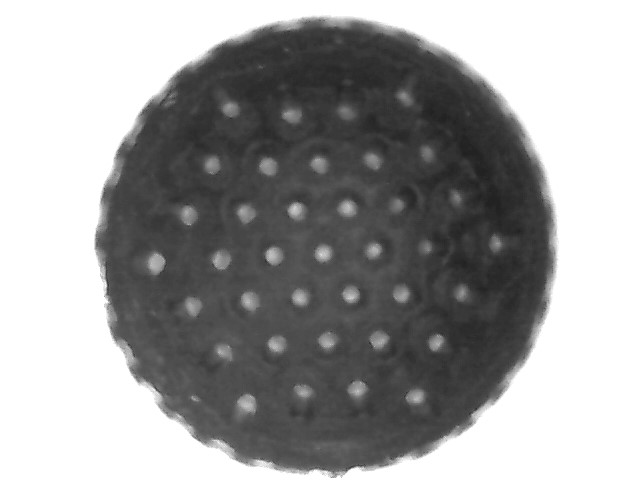} &
                    \includegraphics[width=.47\columnwidth]{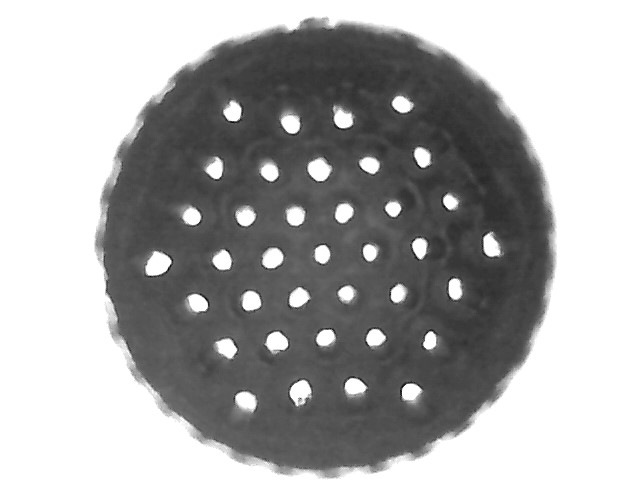} \\[5pt]
                \end{tabular}

              \caption{A comparison of normal pins and pins painted with phosphorescent paint. The ambient light is enough for the painted pins to be clearly visible  without previous exposure to bright light, but not enough for the normal pins. The tips are shown here without gel and lens.}
              \label{lighting-flouresence}
        \end{figure}

        By painting the pins with a phosphorescent paint, specifically an acrylic paint base mixed with ``LIT - the world’s glowiest glow pigment by Stuart Semple", which happens to also fluoresce, the need for continual on-board lighting is reduced. \autoref{lighting-flouresence} demonstrates the difference in pin contrast achieved with application of the paint.
        
    \subsection{Edge-following Experiment}
    
        \begin{figure}[b] 
            \centering
          
            \begin{tabular}[b]{@{}ccc@{}}
                {\bf \footnotesize Data Collection Arc} & {\bf \footnotesize Directly on Edge} & {\bf \footnotesize  Safe Offset from Edge} \\[0pt]
            \end{tabular}
            \includegraphics[width=.98\columnwidth]{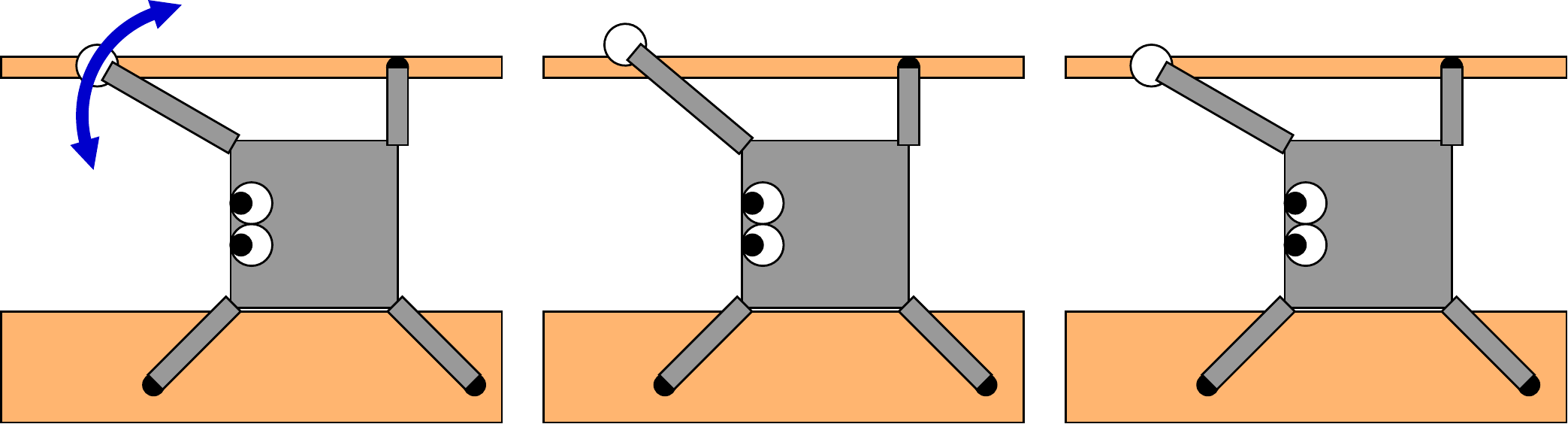} 

            \caption{Diagram showing the key robot poses from above during beam walking experiment as robot attempts to move from right to left. Left: Arrow indicates the angles over which the robot collects data to add the models. Center: Stance when robot has successfully located the edge. Right: Stance when robot places its foot a safe distance from the found edge.}
            \label{pose-diag}
        \end{figure}
    
        The robot is placed on two raised beams of wood, one just wider than the tactile foot (28mm), the other wide enough for the sensorless feet to never fall off. The beams are tall enough that the foot cannot touch the ground when tapping either side of the beam. The aim is for the robot to track the outer edge of the narrow beam with the tactile foot, placing the foot a set distance from the edge to ensure a stable foothold. With the tactile sensing turned off, the robot is commanded to walk in a straight line, however, it always falls off the narrow beam due to motor control inaccuracies and subtle changes in beam height (the wooden beams are not perfectly flat, nor held perfectly parallel, reflecting the imperfections often encountered in the real world). Hence this set-up demonstrates the usefulness of a tactile foot.
        
        In order to implement this experiment the high-level logic in \autoref{logic} was used to control the robot.

        \begin{algorithm}
            \SetAlgoLined
            \tcp{Initialise}
            Tap in an arc with front right foot\;
            Initialise models and find edge\;
            Plant foot at safe distance from edge\;
            Turn body to align with front right foot\;
            Walk forward until right leg lifted again\;
            
            \tcp{Main loop}
            \While{current iteration $<$ max iterations}{
                Place front right foot forward and tap\;
                Estimate location of edge\;
                Move foot to edge\;
                \If{edge not directly under foot}{
                    Collect another arc of taps to add to model\;
                    Find actual location of edge\;
                }
                Place foot at safe distance from found edge\;
                Turn body to align with front right foot\;
                Walk forward until right leg lifted again\;
            }
            \caption{High-level Contour Following\label{logic}}
        \end{algorithm}
    
        \begin{figure*}[t]
            \centering
            
            {\bf \footnotesize Approximate Motion of front right foot during beam walking}\\
            \includegraphics[trim={2.6cm 0.5cm 2.5cm 0.8cm}, clip, width=.98\textwidth]{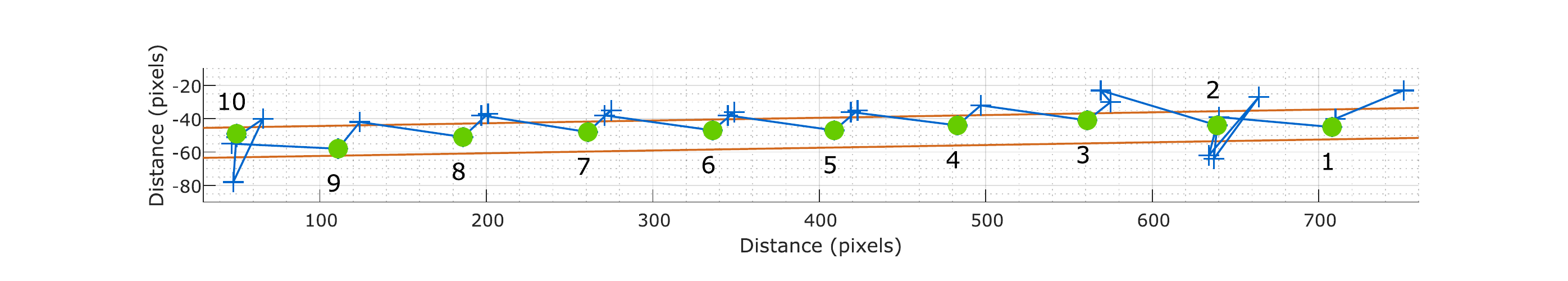}\\
            
            \caption{The key locations of the front right foot throughout the experiment, relative to the beam (brown outline). The blue crosses indicate all taps and footholds (only the limits of data collection arcs are shown), the blue line connects these points to indicate order of positions, and the green circles indicate the chosen footholds (labelled from 1 to 10). }
            \label{plot}
        \end{figure*}

    \subsubsection{Online Learning}
        As described by Stone et al.~\cite{stone2019learning}, online learning can be used to drastically decrease the training time needed to build a model to accurately interpret tactile stimuli for edge following. The experiment in this paper resembles that previous setup, with the robot arm replaced with a walking robot, and so similar principles can be applied here.  To adapt the methods to function on a walking robot, the search line is an arc rather than a straight line as this is the only way of moving that keeps the same angle of incidence between the foot and ground. This was coincidentally posed as an improvement and solution in the original work for loss of edge around tight corners. As such, the angle of the robots hip joint is used as a proxy for displacement.
        
        Instead of collecting a large dataset with the sensor in many orientations on one stimuli, as would be done in off-line learning, the robot simply collects data as and when it is needed, as determined by the accuracy of the model predictions. Elaborating on \autoref{logic}, as an initial model the robot collects an arc of data in its original pose (see \autoref{pose-diag}). This data is then used to initialise a Gaussian Process Latent Variable Modelling (GP-LVM) model, with the aid of a dissimilarity measure to align the unknown arc displacement with a reference tap to give labelled data (we refer to ref.~\cite{stone2019learning} for details of this method). Having found the edge the robot is able to orientate its body to walk at this offset from the original pose. 
        
        To ensure reliable foot placement with proceeding taps the robot taps twice, firstly where the edge would be if the robot walked perfectly in line with the beam. This first tap is used to estimate the displacement to the edge and the second tap is taken after moving by this displacement and should with an accurate model be directly on the edge, i.e. have a predicted displacement of 0\textdegree\hspace{0.1pt}. If the prediction is not within a set tolerance ($\pm$3\textdegree) of this  the model must be wrong and so more data is collected to correct it, otherwise the robot continues without collecting more data.  
   
        By using these online methods, the sensor is trained during the current task and therefore models consist of data that reflect  the  current task. Sensor characterisation was also  not needed beforehand for a specific sensor, meaning a tip can easily be replaced at any time without needing to retrain sensor models. The only up-front data needed was an arc of evenly spaced taps used to select a point to use as a reference for the edge to follow. 
    
        \begin{figure}[b]
            \centering
            {\bf  Pin Detection in Different Ambient Light Conditions}\\
            \vspace{2pt}
            
            \includegraphics[trim={5cm 0.55cm 6.5cm 1.1cm}, clip,width=.24\columnwidth]{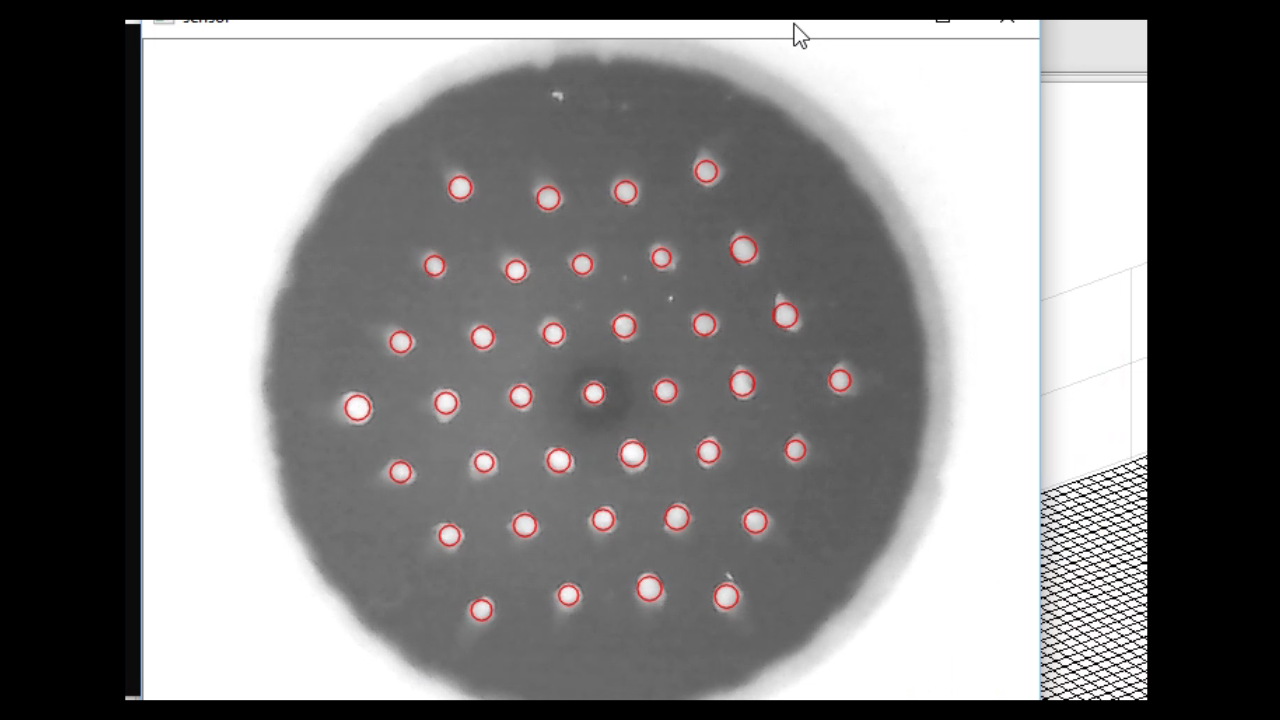} 
            \includegraphics[trim={5cm 0.55cm 6.5cm 1.1cm}, clip,width=.24\columnwidth]{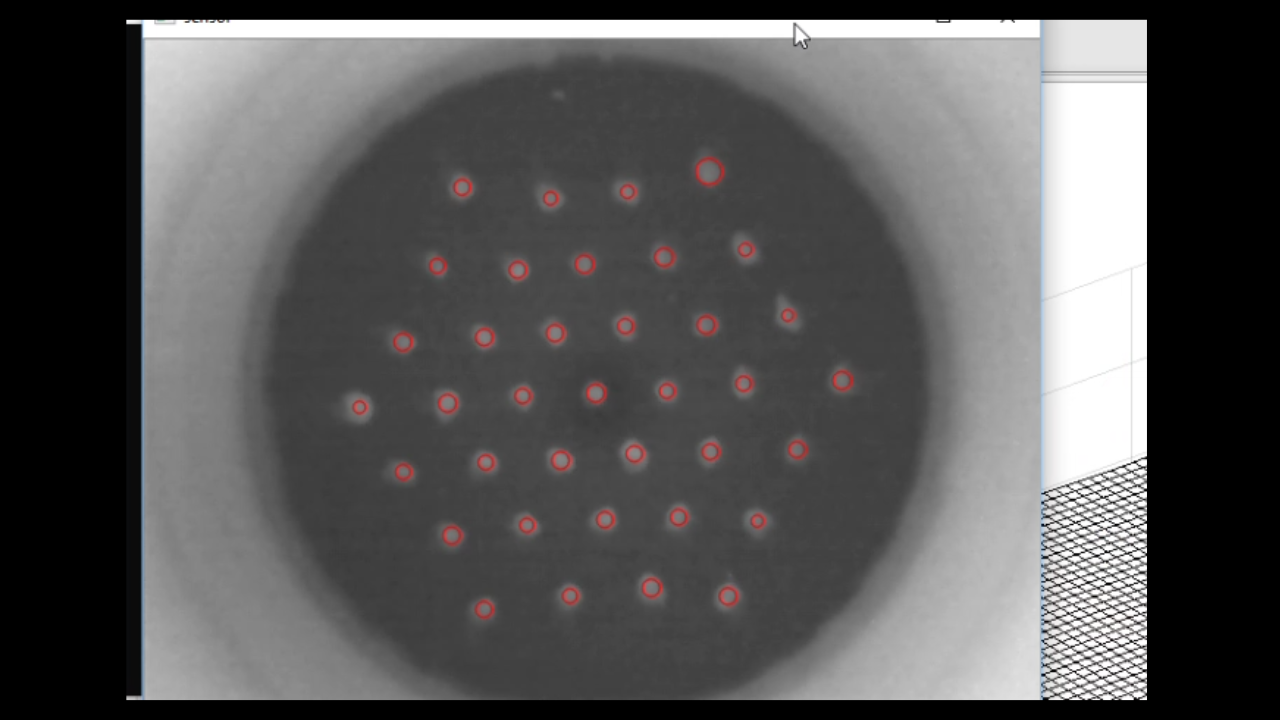}
            \includegraphics[trim={5cm 0.55cm 6.5cm 1.1cm}, clip,width=.24\columnwidth]{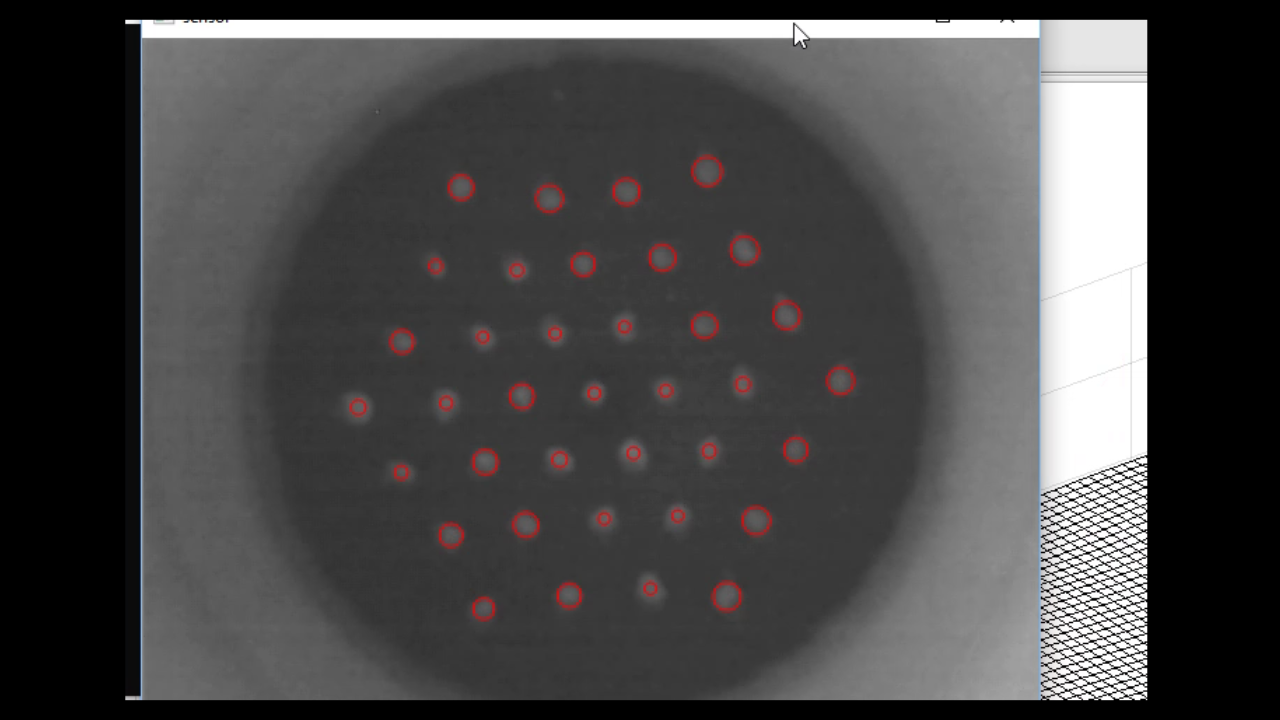} 
            \includegraphics[trim={5cm 0.55cm 6.5cm 1.1cm}, clip,width=.24\columnwidth]{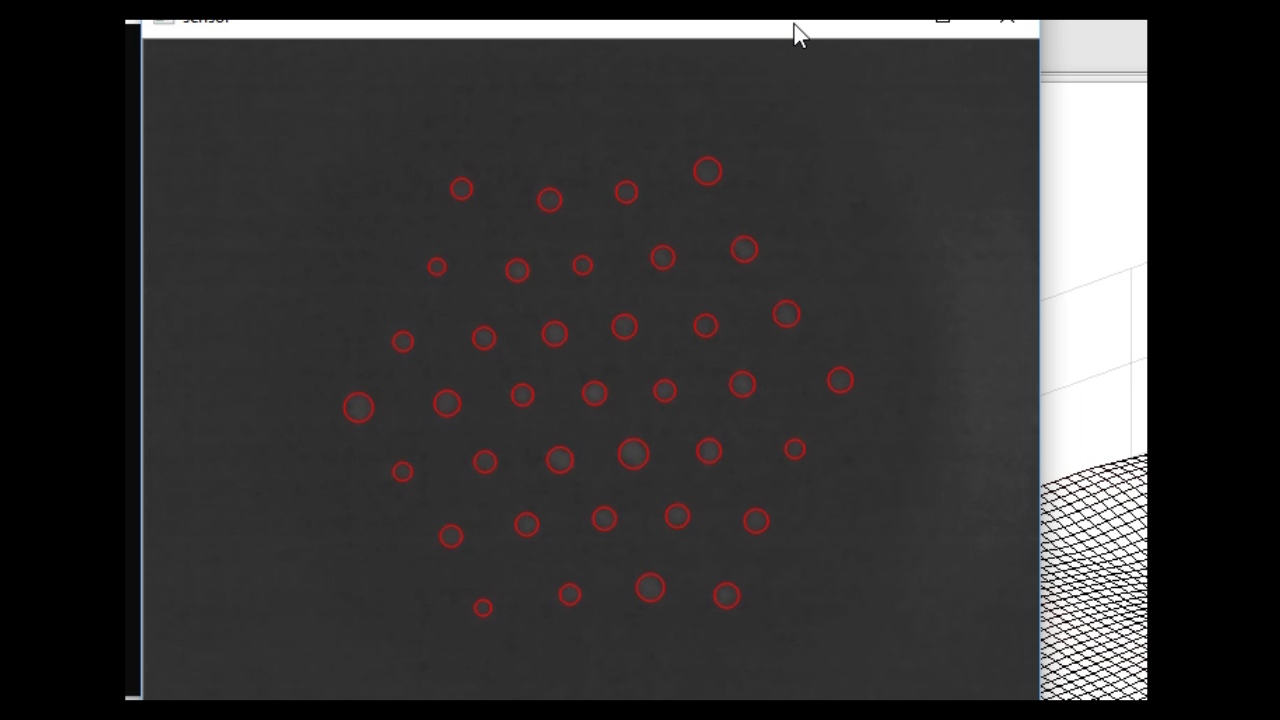} \\[0pt]
            \begin{tabular}[b]{@{}cccc@{}}
                {\bf \footnotesize  Ambient Light } & {\bf \small  \hspace{80pt}   } & {\bf \small    \hspace{2pt}} & {\bf \footnotesize No Ambient Light}
            \end{tabular}
            
            \caption{During bright glow phase, how pins look under different ambient light levels. The red circles are the locations of the pins according to computer vision - as you can see the pins are successfully identified under different ambient light conditions. }
            \label{lighting-differing}
        \end{figure}

\section{Results}\label{results}

    \subsection{Initial Validation of Tactile Foot}
        Charging the pins with internal LEDs for around 10 seconds allows the pins to be detected for 30 to 40 seconds with no ambient light at all, allowing  LED use  to be drastically reduced. As can be seen in \autoref{lighting-differing} the pins remain detectable using computer vision. Minimal modification was done to the original pin detection methods. Occasionally in low-light levels the black background was mistaken for pins. In principle, this could be resolved by tuning the computer-vision techniques.
        
        Without exposure to bright light, however, the paint fluoresces bright enough from only ambient light diffused through the foot cone that the pins can be detected without need for internal lighting (see \autoref{lighting-flouresence}). As such internal LEDs can be permanently switched off in most settings, as is the case in the other experiments here.
        
        Leveraging these two factors, this simpler to manufacture sensor remains functional in all settings at a theoretically reduced power consumption, and with more reliable pin detection (as there are no LED lights to be erroneously detected as pins).

    \subsection{Narrow Beam Walking}
    
        \begin{figure}[t]
            \centering
            
            {\bf \footnotesize Overlay of Locations from Video}\\
            \includegraphics[width=.98\columnwidth]{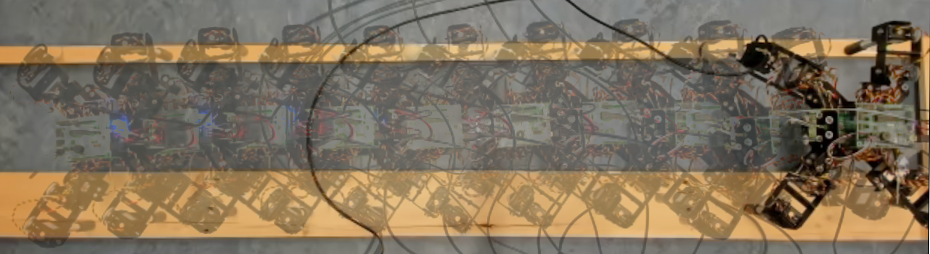}\\
            
            \caption{Overlay of all the locations selected to place foot as the robot walks from right to left. As you can see, the robot kept a straight line along the beam and did not fall off. }
            \label{blur}
        \end{figure}
      
        \begin{figure}[t]
            \centering
            
            {\bf \footnotesize Foothold Displacement from beam center}
            \includegraphics[trim={0.5cm 0cm 1.3cm 0cm}, clip, width=.98\columnwidth]{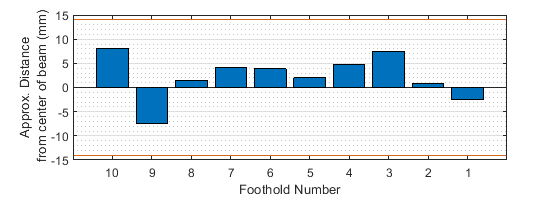}
            
            \caption{Approximate displacement from the center of the beam of the center of the foot at each foothold. The brown lines indicate the edge of the beam, where exactly half of the sensor would be on and off the beam. Anything between these two lines is a stable foothold.}
            \label{bar}
        \end{figure}
 
        \begin{figure}[h]
            \centering
            {\bf \footnotesize Representation of GP-LVM Model}\\
            \vspace{5pt}
            \includegraphics[trim={0cm 0.5cm 0cm 0.9cm}, clip, width=.9\columnwidth]{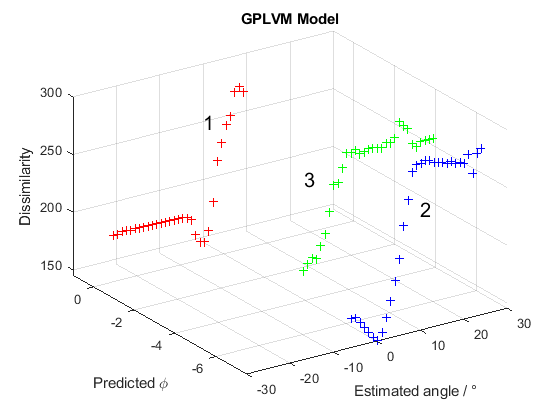}
            \caption{Representation of the GP-LVM model. Dissimilarity shows how close the data resembles the reference tap (the edge), with 0\textdegree~being the closest match with the reference tap. In the actual model dissimilarity is not used, as it directly uses the $x,y$ coordinates of each pin.  }
            \label{gplvm}
        \end{figure}
    
        As can be seen in \autoref{blur} and the accompanying video, the robot was able to walk along the narrow beam without deviating or falling off. Based on data extracted from the video recording of the experiment, the foot was on average (absolute mean) within 4mm of the center of the beam, with overall range within 8mm of the center of the beam. As long as the center of the foot remains within the width of beam the foothold is considered safe and therefore all footholds were well within the safe limits (even the largest deviation from the center of the beam was still 6mm below this limit). More accurate placement of the foot could be obtained by reducing the tolerance, whereby data would be collected more often. 
        
        Furthermore the robot was able to complete this task using very little data. As can be seen in \autoref{plot} and \autoref{gplvm}, the model consisted of only 3 arcs of data, two at the start and one at the end of the experiment; this is a total of 91 taps used to train the model.
    
        Note also that the last tap arc was not so accurate due to change in height of beam object. This was perceived as changes in edge displacement  due to the hemispherical-shaped tip. This may be solved by allowing the model to train on different heights. 
        Inaccuracies may also be caused by the uneven distribution of taps about 0\textdegree\hspace{0.5pt}, especially on the last data collection arc where it is unclear if the true minimum dissimilarity was found. This could be improved by simply extending the arc, or introducing a search routine to find the edge so data can be collected evenly both sides of it. 
        
        The accompanying video shows the robot walking slowly with it pausing after each prediction -- this is overly cautious and can be easily removed to significantly speed up the robot.
    
    \subsection{Following a Curved Edge}
\begin{figure}[bt]
            \centering

            \includegraphics[width=.98\columnwidth]{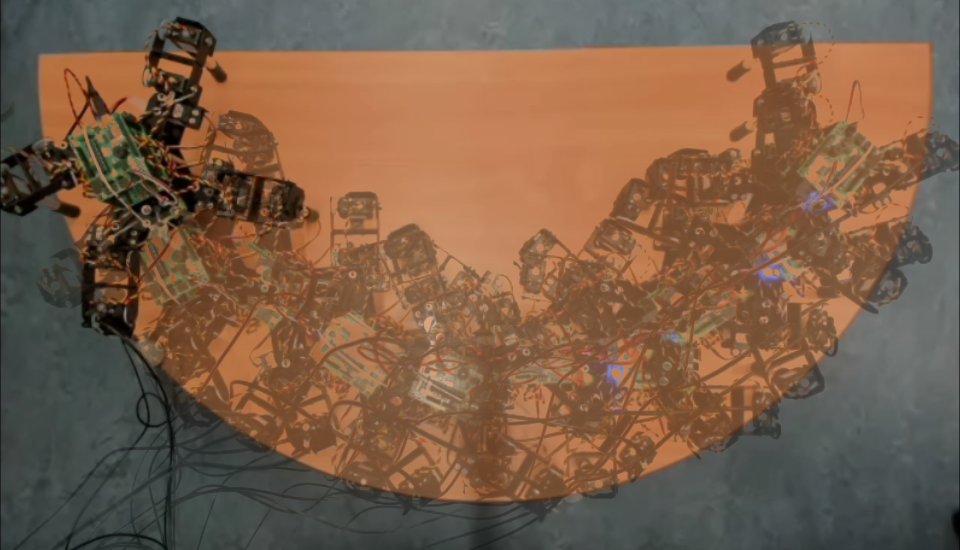}

            \caption{Overlay of all the locations selected to place foot as the robot moves anti-clockwise around a semi-circular table. The robot correctly sensed the edge along the entire semi-circular section of the table.}
            \label{blur-table}
        \end{figure}

        To move around a curved edge, the online learning methods needed modification to enable the robot to sense large deviations of the edge, as encountered with the 59cm radius table in \autoref{blur-table}. An additional search algorithm was introduced to allow the robot to find the edge when it has moved outside of the sensor's range, which happens because of the relatively-tight curvature of the table edge and the relatively small footprint of the sensor. 
        
        With these modifications, the robot was able to detect the edge of the table and place the front right foot on a safe place on the table, along the entire semi-circular edge. The turning algorithm employed, which worked at small angles, was no longer sufficient at larger angles and therefore the robot did occasionally slip off the edge while trying to turn. In these initial experiments, we solved this by replacing the robot on the table with the front right foot in the same location as previously. This problem is not due to the tactile sensing or online learning, but solely due to the control algorithm that gives the turning behaviour; we expect these can be improved to circumvent this issue. Importantly, the robot was able to sense the edge and place its foot in a safe place.
    
\section{Discussion \& Conclusion}
    This paper has described the development of a modular, easily manufactured, inexpensive and simple biomimetic optical tactile foot, and demonstrated its use on a small walking robot. With the use of this tactile sensor the robot was able to detect the edge of a beam and a curved semi-circular table edge, and place its foot a safe position relative to this edge, allowing it to safely traverse the hazardous terrain. These simple experiments have shown the usefulness of tactile feet for walking robots. In addition to the these and other benefits discussed in this paper, the same sensor has been used for series of trials over at least 12 hours, yet shows no signs of degradation. This is a major improvement over the sensor proposed by Shill et al. \cite{Shill2015} which, while showing a tactile foot can be useful for texture classification, lasted only 8 minutes (53 minutes with degradation compensation). This durability shows promise for future applications of robotic feet.
    
    Online learning was used to efficiently perceive tactile stimuli during the experiment, based on methods proposed by Stone et al.~\cite{stone2019learning} for data-efficient contour following with a TacTip mounted on a robot arm. This efficiency allowed quick and easy setup of experiments across platforms, with no need for time-consuming data collection to retrain the sensor on the leg. These low-data principles could be extended to other tasks with the design of efficient data collection policies. This means once data collection policies are discovered to enable online learning, a task developed on one platform can quickly be transferred to any other platform e.g. from arms to legs, between different walking robots or even between different morphology sensors. In addition, as learning is never stopped, a walking robot traversing challenging terrains will be constantly learning and adapting to its environment, not limited by an initial training set trained on limited stimuli with offline learning. This has the potential to increase the robustness of walking robots across a variety of terrains.
    
    This study has shown how tactile feet are a feasible and valuable contribution to the sensory systems of  walking robots.
    This work has only explored one of the many modes of feedback available from the TacTip sensor, from only a single foot, and combined this with a simplistic control strategy. Combination with more feedback modes, more tactile feet and more intelligent control would enable significantly more complex environments to be traversed. Future modes of tactile feedback that could be explored include texture classification, slip detection, angle of surface slope, 3D edge detection and surface compliance.
    
    The benefits of tactile feet are not limited to this setup as the design can be easily adapted to any model and weight of robot. The limiting factor in the current design is the rigidity of the gel inside the tip; this gel can be replaced with a stiffer material, as in  Elkington et al.~\cite{Elkington2020} who considered modified tips capable of withstanding forces of up to 400N. Their limiting factor was the strength of the 3D printed casing rather than the tip gel; therefore, with a stronger case material the sensor could  withstand much larger forces. This would enable this work to be carried over to other larger platforms with much greater forces through the feet, to enable them to traverse more challenging terrains.




\section*{ACKNOWLEDGMENT}
We thank Andrew Stinchcombe and Patrick Brinson for all their help, and the many others in the Tactile Robotics group at Bristol Robotics Laboratory.




\bibliographystyle{IEEEtran}
\bibliography{IEEEabrv, main}

\end{document}